\documentclass[sigconf]{acmart}

\renewcommand\footnotetextcopyrightpermission[1]{}
\usepackage{bbding}

\usepackage{amssymb}
\usepackage{multirow}

\usepackage{hyperref}

\author{Zhuopeng Li}
\affiliation{%
   \institution{Zhejiang University}
   \country{China}}
\email{lizhuopeng@zju.edu.cn}

\author{Lu Li}
\affiliation{%
   \institution{Zhejiang University}
   \country{China}}
   
   \author{Zeyu Ma}
\affiliation{%
   \institution{Alibaba Group}
   \country{China}}     
   
   \author{Ping Zhang}
\affiliation{%
   \institution{Alibaba Group}
   \country{China}}    
   
 \author{Junbo Chen}
\affiliation{%
   \institution{Alibaba Group}
   \country{China}}   
   
 \author{Jianke Zhu}
\affiliation{%
   \institution{Zhejiang University}
   \country{China}}

\begin{document}

\title{READ: Large-Scale Neural Scene Rendering for \\ Autonomous Driving}


\begin{abstract}
Synthesizing free-view photo-realistic images is an important task in multimedia. With the development of advanced driver assistance systems~(ADAS) and their applications in autonomous vehicles, experimenting with different scenarios becomes a challenge. Although the photo-realistic street scenes can be synthesized by image-to-image translation methods, which cannot produce coherent scenes due to the lack of 3D information. In this paper, a large-scale neural rendering method is proposed to synthesize the autonomous driving scene~(READ), which makes it possible to synthesize large-scale driving scenarios on a PC through a variety of sampling schemes. In order to represent driving scenarios, we propose an $\omega-net$ rendering network to learn neural descriptors from sparse point clouds. Our model can not only synthesize realistic driving scenes but also stitch and edit driving scenes. Experiments show that our model performs well in large-scale driving scenarios. \\

\href{https://github.com/JOP-Lee/READ-Large-Scale-Neural-Scene-Rendering-for-Autonomous-Driving}{Github:  https://github.com/JOP-Lee/READ-Large-Scale-Neural-Scene-Rendering-for-Autonomous-Driving}
\end{abstract}



\begin{teaserfigure}
\centering
  \includegraphics[width=\linewidth]{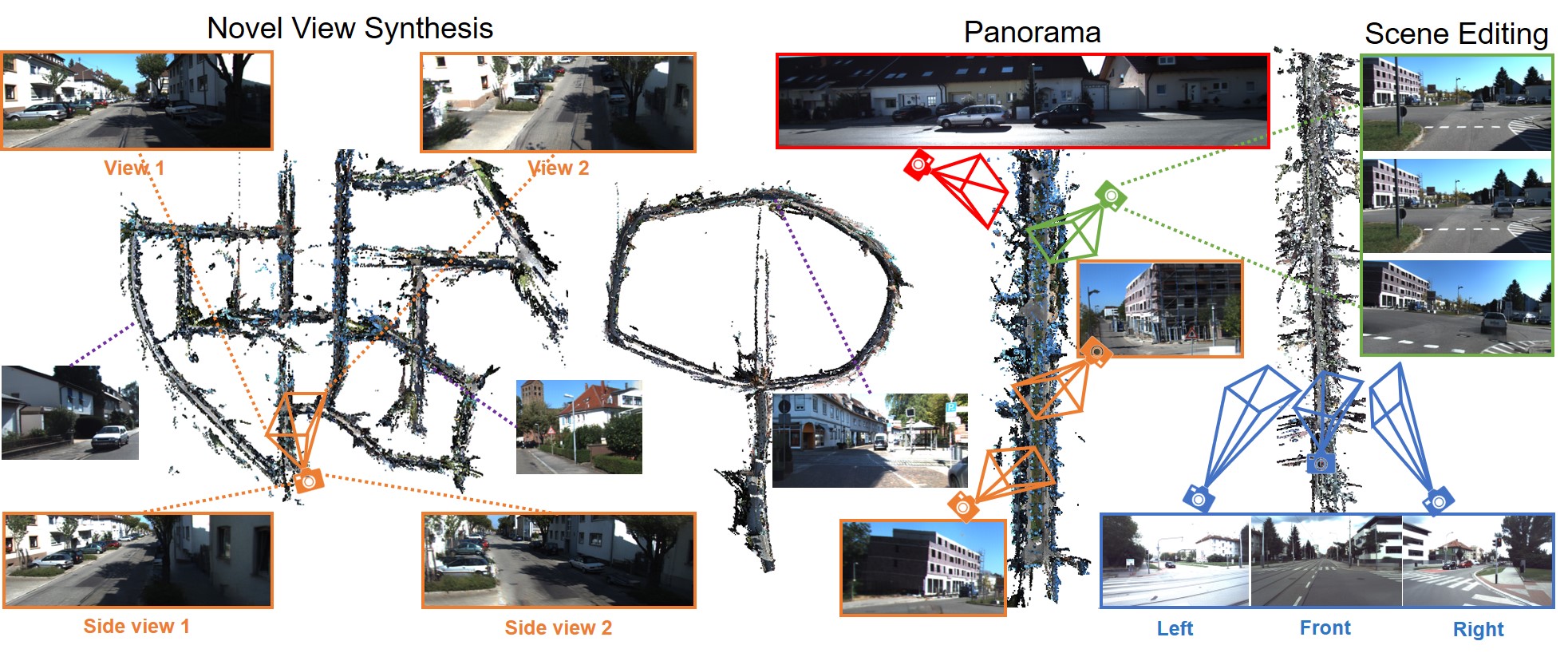}  
  \caption{Given the input point clouds, our Autonomous Driving scene Render~(READ) synthesizes photo-realistic driving scenes from different views, which is able to provide rich data for autonomous driving rather than images with a single view.}
  \label{fig:label1}
\end{teaserfigure}

\maketitle

\section{Introduction}



Synthesizing free-view photo-realistic images is an important task in multimedia~\cite{chen2019multimedia}. Especially, the synthetic large-scale street views are essential to a series of real-world applications, including autonomous driving~\cite{li2019event,kim2013parallel}, robot simulation~\cite{dosovitskiy2017carla,wiriyathammabhum2016computer}, object detection~\cite{zhang2021auto,zhang2021depth,he2021end}, and image segmentation~\cite{ying2021srnet,gao2020lab2pix,tang2020dual}.
As illustrated in Fig.~\ref{fig:label1}, the objective of neural scene rendering is to synthesize the 3D scene from a moving camera, where the user can browse the street scenery from different views and conduct automatic driving simulation experiments. In addition, this can generate multi-view images to provide data for multimedia tasks. 



With the development of autonomous driving, it is challenging to conduct experiments in various driving scenarios. Due to the complicated geographic locations, varying surroundings, and road conditions, it is usually difficult to simulate outdoor environments. Additionally, it is hard to model some unexpected traffic scenarios, such as car accidents, where the simulators can help to reduce the reality gap. However, the data generated by the widely used simulator like CARLA~\cite{dosovitskiy2017carla} is far different from real world scenes using the conventional rendering pipeline. 

The image-to-image translation-based methods~\cite{gao2020lab2pix,tang2020dual,isola2017image, tang2020local} synthesize the street views with semantic labels by learning the mapping between source images and targets. Despite of generating the encouraging street scene, there exist some large artifacts and incoherent textures. Moreover, the synthesized image has only a single view that cannot provide the rich multi-view traffic conditions for autonomous vehicles. This hinders them from a large number of real world applications.

 Recently, Neural Radiance Field~(NeRF) based methods~\cite{zhang2021ners,niemeyer2021giraffe,mildenhall2020NeRF,wang2021ibrnet} achieve the promising results in synthesizing the photo-realistic scenes with multi-view. As suggested in~\cite{kangle2021dsnerf}, they cannot produce reasonable results with only few input views, which typically happens in the driving scenario with the objects appearing in only a few frames. Moreover, the NeRF-based methods mainly render either the interiors or objects. They have difficulty synthesizing the large scale driving scenes with the complicated environment, where the large artifacts occur in the closed-up views and surroundings. To tackle this problem, NeRFW~\cite{martin2021nerf} makes use of the additional depth and segmentation annotations to synthesize an outdoor building, which takes about two days with 8 GPU devices. Such a long reconstruction time is mainly due to the unnecessary sampling of the vast spaces. 

 Unlike the NeRF-based methods that purely depend on per-scene fitting, the neural rendering approaches~\cite{thies2019deferred,wang2021ibutter,wu2021imitating} can be effectively initialized via neural textures, which are stored as maps on top of the 3D mesh proxy. Similarly, NPBG~\cite{aliev2020neural} learns neural descriptors from a raw point cloud to encode the local geometry and appearance, which avoids sample rays in empty scene space by the classical point clouds reflecting the geometry of the scene in real world. Moreover, ADOP~\cite{ruckert2021adop} improves NPBG by adding a differentiable camera model with a tone mapper, which introduces the formulation to better approximate the spatial gradient of pixel rasterization. In general, the point-based neural rendering method can synthesize a larger scene with fewer captured images by initializing the scene through three-dimensional point cloud information. Although neural rendering-based methods can synthesize the photo-realistic novel views in both indoor and outdoor scenes, it is still very challenging to deal with the large-scale driving scenarios due to the limitations on model capacity, as well as constraints on memory and computation. Additionally, it is difficult to render the photo-realistic views with rich buildings, lanes, and road signs, where the sparse point cloud data obtained from the few input images usually contain lots of holes.
 

In this paper, we propose an effective neural scene rendering approach, which makes it possible to synthesize the large-scale driving scenarios through efficient Monte Carlo sampling, screening of large-scale point clouds, and patch sampling. It is worth mentioning that our method synthesized large-scale driving scenarios with an average of two days of training on a PC with two RTX2070 GPUs. This greatly reduces the computational cost so that large-scale scene rendering can be achieved on affordable hardware. For sparse point clouds, we fill in the missing areas of point clouds by multi-scale feature fusion. To synthesize photo-realistic driving scene from sparse point clouds, we propose an $\omega-net$ network to filter neural descriptors through basic gate modules and fuse features of the same scale and different scales with different strategies. Through $\omega-net$, our model can not only synthesize the realistic scenes but also edit and stitch scenes via neural descriptors. Moreover, we are able to update the specific areas and stitch them together with the original scene. Scene editing can be used to synthesize the diverse driving scene data from different views even for traffic emergencies.

The main contributions of this paper are summarized as: 1) Based on our neural rendering engine~(READ), a large-scale driving simulation environment is constructed to generate realistic data for advanced driver assistance systems; 2) $\omega-net$ network is proposed to obtain a more realistic and detailed driving scenario, where multiple sampling strategies are introduced to enable synthesize the large-scale driving scenes; 3) Experiments on the KITTI benchmark~\cite{geiger2012we} and Brno Urban dataset~\cite{ligocki2020brno} show the good qualitative and quantitative results, where the driving scenes can be edited and stitched so as to synthesize larger and more diverse driving data.


\section{Related Work}
\subsection{Image-to-image Translation}
Many researchers~\cite{gao2020lab2pix,tang2020dual,isola2017image,tang2020local,richter2021enhancing} employ image-to-image translation technique to synthesize photo-realistic street scenes. Gao et al.~\cite{gao2020lab2pix} propose an unsupervised GAN-based framework, which adaptively synthesizes images from segmentation labels by considering the specific attributes of the task from segmentation labels to image synthesis. As in~\cite{gao2020lab2pix}, Tang et al.~\cite{tang2020dual} present a dual-attention GAN that synthesizes the photo-realistic and semantically consistent images with fine detail from input layouts without the additional training overhead. To enhance the fidelity of images in the game, Richter et al.~\cite{richter2021enhancing} use the G-buffers generated in the rendering process of the game engine as the additional input signal to train the convolutional neural network, which is able to eliminate the disharmonious and unreasonable illusions generated by previous deep learning methods. Although the image-to-image translation method can synthesize the realistic street scene, it still cannot guarantee coherence in the scene transformation. Moreover, it can only synthesize the scene from a single view, whose results are far different from the real scene in terms of texture details.

\subsection{Novel View Synthesis}
Neural Radiance Fields~\cite{mildenhall2020NeRF} become an important breakthrough for the novel view synthesis task, which is proposed to use a fully connected network of entire scenes optimized by differentiable volume rendering. Recently, there have been many variations of this method to render different objects, such as human~\cite{Humannerf}, car~\cite{niemeyer2021giraffe}, interior scene~\cite{wang2021ibrnet}, and building~\cite{martin2021nerf}. However, NeRF-based methods depend on per-scene fitting, it is hard to fit a large-scale driving scenario. NeRFW~\cite{martin2021nerf} combines appearance embedding and decomposition of transient and static elements through uncertainty fields. Unfortunately, dynamic objects are ignored, which may lead to occlusions in the static scene. Moreover, synthesizing scenes as large as street view requires huge computing resources, which cannot be rendered in real-time. 

 \begin{figure*}[t]
  \centering
  \includegraphics[width=\linewidth]{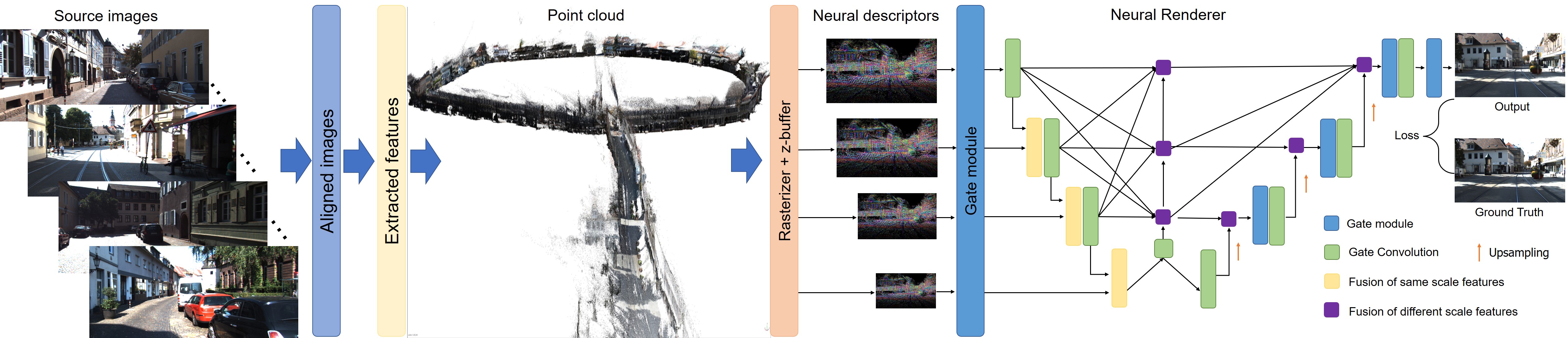}
  \caption{Overview of our proposed large-scale neural scene Render (READ) for Autonomous Driving. The input image is firstly aligned, and then the point cloud of the scene is obtained by matching feature points and dense construction. We rasterize points at several resolutions. Given the point cloud $P$, the learnable neural descriptor $D$, and the camera parameter $C$, our presented $\omega-net$ rendering network synthesizes realistic driving scenes by filtering neural descriptors learned from the data and fusing the features from the same scale and different scales.}
\label{fig:label2}
\end{figure*}

\subsection{Scene Synthesis by Neural Rendering}
Meshry et al.~\cite{meshry2019neural} take a latent appearance vector and a semantic mask of the transient object's location as input, which render the scene's points into the deep frame buffer and learn these initial render mappings to the real photo. This requires a lot of semantic annotation and ignores the transient objects. By combining the traditional graphics pipelines with learnable components, Thies et al.~\cite{thies2019deferred} introduce a new image composition paradigm, named Deferred Neural Rendering, where feature mapping of the target image is learned from UV-map through neural texture. Despite the promising results, it is time-consuming to obtain explicit surfaces with good quality from the point cloud.

The point-based neural rendering method employs point clouds as input to learn the scene representation. NPBG~\cite{aliev2020neural} encodes the local geometric shapes and appearance by learning neural descriptors, which synthesizes high quality novel indoor views from point clouds. TRANSPR~\cite{kolos2020transpr} extends NPBG by augmenting point descriptors with alpha values and replacing Z-buffer rasterization with ray marching, it is able to synthesize semi-transparent parts of the scene. ADOP~\cite{ruckert2021adop} proposes a point-based differentiable neural rendering, where the parameters of all scenarios are optimized by designing the stages of the pipeline to be differentiable.

\section{Large-Scale Neural Scene Render}\label{sec:LNSR}
Our proposed Neural Scene Render approach aims to synthesize the photo-realistic images from an arbitrary camera viewpoint by representing the driving scenes with point clouds. In this section, we first outline our proposed method. Secondly, multi sampling strategies of sparse point clouds are proposed to reduce the computational cost for large-scale driving scenes. Thirdly, $\omega-net$ is proposed to represent driving scenes with sparse point clouds and synthesize realistic driving scenarios. Finally, the driving scenes are edited and stitched to provide synthetic data for larger and richer driving scenes.

\subsection{Overview}
Given a set of input images for a driving scene and the point cloud $P=\left\{p_{1}, p_{2}, \ldots, p_{N}\right\}$ with known camera parameters, our framework is capable of synthesizing the photo-realistic driving scenes from multiple views, as well as stitching and editing driving scenes. To this end, we propose an end-end large-scale neural scene render that synthesizes realistic images from sparse point clouds. Our framework is divided into the following three parts: rasterization, sampling with the sparse point cloud, $\omega-net$ rendering network. The overview of our proposed framework is illustrated in Fig.~\ref{fig:label2}.


Sparse 3D point cloud $P_{i}$ can be obtained through the classic Structure-from-Motion and Multi-View Stereo pipelines, such as Agisoft Metashape~\cite{Metashape}. Each point $i$ is located at $p_{i}$, which is associated with a neural descriptor vector encoding the local scene content. As in~\cite{aliev2020neural}, each input 3D point in $P$ contains the position $p_{i}=\left\{x_{i}, y_{i}, z_{i}\right\}$, whose appearance feature is extracted by mapping the RGB value of image pixel to its corresponding 3D space. Neural descriptors $D_{i}=\left\{d_{1}, d_{2}, \ldots, d_{N}\right\}$ are calculated from the input point cloud, namely latent vectors representing local geometry and photometric properties. We update these features by propagating gradient to the input so that the features of the neural descriptor can be automatically learned from data. Given the camera's internal and external parameters, we can view the scene from different views by designing 8-dimensional neural descriptors to represent the RGB values. In the rasterization phase, images of size $W \times H$ are captured by pinhole camera $C$, we construct a pyramid of rasterized raw images $\left\{S_{t}\right\}_{t=1}^{T}$ (T=4 in all our experiments), $S_{t}$ has the spatial size of $\frac{W}{2^{t}} \times \frac{H}{2^{t}}$, which is formed by assigning the neural descriptor of the point passing the depth test to each pixel. Then, it is projected onto the pixel under the full projection transformation of the camera. Essentially, the neural descriptors feature $D_{i}$ encodes the local 3D scene content around $p_{i}$. The rendering network expresses a local 3D function $f$ that outputs the specific neural scene description $f(i, x)$ at $x$, modeled by the neural point in its local frame.

 \begin{figure}[h]
  \centering
  \includegraphics[width=\linewidth]{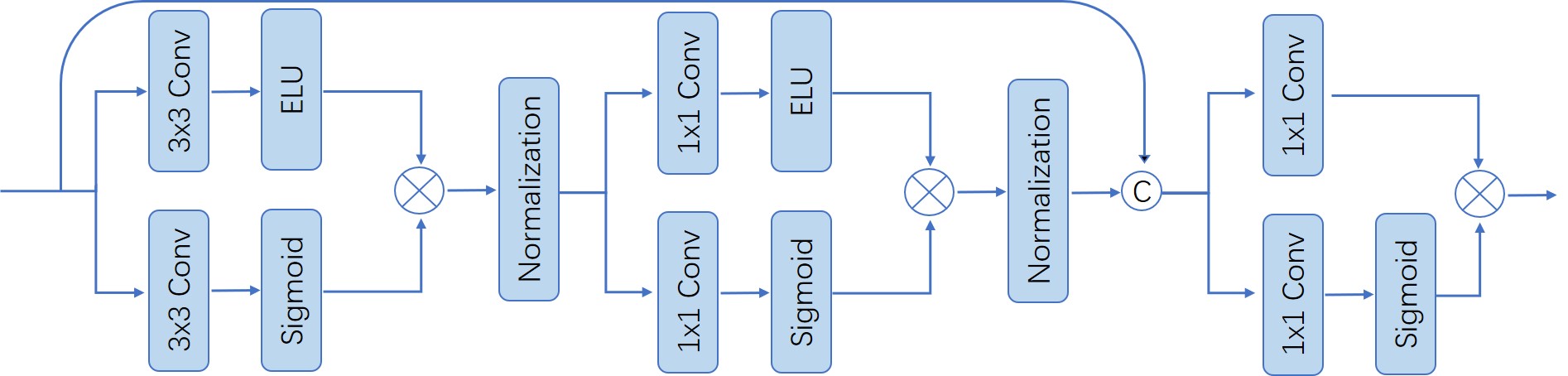}
  \caption{Basic gate module. Neural descriptors learned from sparse point clouds can effectively screen out invalid values.}
  \label{fig:label3}
\end{figure}

\subsection{Sampling with Sparse Point Cloud}
Synthesizing driving scenes with thousands of meters requires enormous computational power. Therefore, the key to our proposed approach is to reduce memory usage and improve training efficiency. Instead of fitting each scene separately, we employ the point clouds generated by the off-the-shelf 3D reconstruction software~\cite{Metashape} to initialize the geometry of the real world scene. As a huge amount of point cloud data consumes a lot of memory, it is still difficult to train. To tackle this critical issue, we take advantage of a sparse sampling strategy to generate sparse point clouds from 1/4 of the originally available pixels, resulting in only 25\% of the total number of point clouds trained.  

\subsubsection{Screen out occluded point clouds}
To avoid updating the descriptors of the occluded points, we approximate the visibility of each point. We use the nearest rasterization scheme by constructing a Z-buffer, reserving only the points with the lowest Z-value at the pixel position. Neural descriptors avoid the points that are far from the camera of the current frame to better synthesize the scene. Thus, the computational cost of calculating occluded point cloud is reduced so that the training efficiency is greatly improved.

\subsubsection{Monte Carlo sampling}

Due to the different distribution of point clouds in the scene, there are abundant point clouds in the area with obvious features. For the area of the sky or dynamic objects, there are fewer corresponding regional point clouds due to the lack of obvious features or fewer feature points. To train effectively, we propose dynamic training strategies, which take advantage of the Monte Carlo method~\cite{shapiro2003monte} to sample a large amount of driving scene data. For the image set $S_e$ in the training phase $e$,
\begin{equation}
  S_{e}^{*} \leftarrow \arg \mathrm{Top}_{n} Q(I_{e}).
\end{equation}
$Q(I_{e})$ is the synthetic quality of image $I_{e}$, which is calculated by perceptual loss~\cite{johnson2016perceptual} in our task. We employ the $\mathrm{Top}_{n}$ samples with the worst performance at each phase as training data. Through the training strategy of dynamic sampling, the model strengthens to learn the sparse region of the point cloud so that the overall training time is reduced.
  
\subsubsection{Patch sampling} 
Image resolution also plays a very important role in memory usage. To this end, we randomly divide the whole image into multiple patches through a sampling strategy, which can select the random patches with the size of $w\times h$ according to the available GPU memory size. It is worth mentioning that the proportion of pixels in patch $w\times h$, to the whole image $W\times H$ is less than 15\% in our task. Given the intrinsic matrix $K$ as below  
$$
K=\left[\begin{array}{ccc}
f_{x} & 0 & c_{x} \\
0 & f_{y} & c_{y} \\
0 & 0 & 1
\end{array}\right]
$$
whee $f_x$ and $f_y$ represent the focal lengths of the $x$ and $y$ axes, respectively. $(c_{x}, c_{y})$ is the position of the principal point with respect to the image plane. 

For each image $I$ in $S_e$, the patch set $S_p$ is obtained by the following strategy to ensure that all areas in the scene can be trained:
\begin{equation}
\begin{gathered}
S_{p}(x, y) \leftarrow \begin{cases}\left(\alpha \cdot f_{x}, \alpha \cdot f_{y}\right) \\
\left(c_{x}+x_{\Lambda}, c_{y}+y_{\Lambda}\right)\end{cases}  
\end{gathered}
\end{equation}
where $\alpha$ is zoom ratio. It shifts the patch $(x_{\Delta},y_{\Delta})$ to enhance the synthetic quality of the scene from different views.


\subsection{ $\omega-net$}
The point cloud, especially ones from external reconstruction methods (e.g., Metashape~\cite{Metashape} or COLMAP~\cite{schonberger2016structure}), often has holes and outliers that degrade the rendering quality. Motivated by MIMO-UNet~\cite{cho2021rethinking}, rendering network $\omega-net$ is proposed to synthesize the novel view from sparse point clouds, which consists of three parts.
  
Given the sparse point cloud $P$, the purpose of the rendering network is to learn reliable neural descriptors to represent scenes. However, neural descriptors learned from point clouds still have holes. To deal with this problem, we design a basic gate module to filter the neural descriptors of different scales, as shown in Fig.~\ref{fig:label3}.
 
By taking into consideration the efficiency, we firstly employ $3 \times 3$ convolution layers $CONV$ to extract the feature of neural descriptor $D_{i}$. A mask is learned by the sigmoid function to filter the invalid values in the neural descriptor. The output is the value range of (0,1), which represents the importance of features in the neural descriptor. To improve the learning efficiency, we employ the ELU activation function for the neural descriptor. $\otimes$ denotes the element-wise multiplication. We concatenate ($\copyright$) the initial feature with the filtered one as a new feature. Finally, we use an additional $1\times1$ convolution layer to further refine the concatenated features. In addition, Gate convolution ~\cite{yu2019free} is introduced to re-filter the fused features.



\begin{figure}[h]
  \centering
  \includegraphics[width=\linewidth]{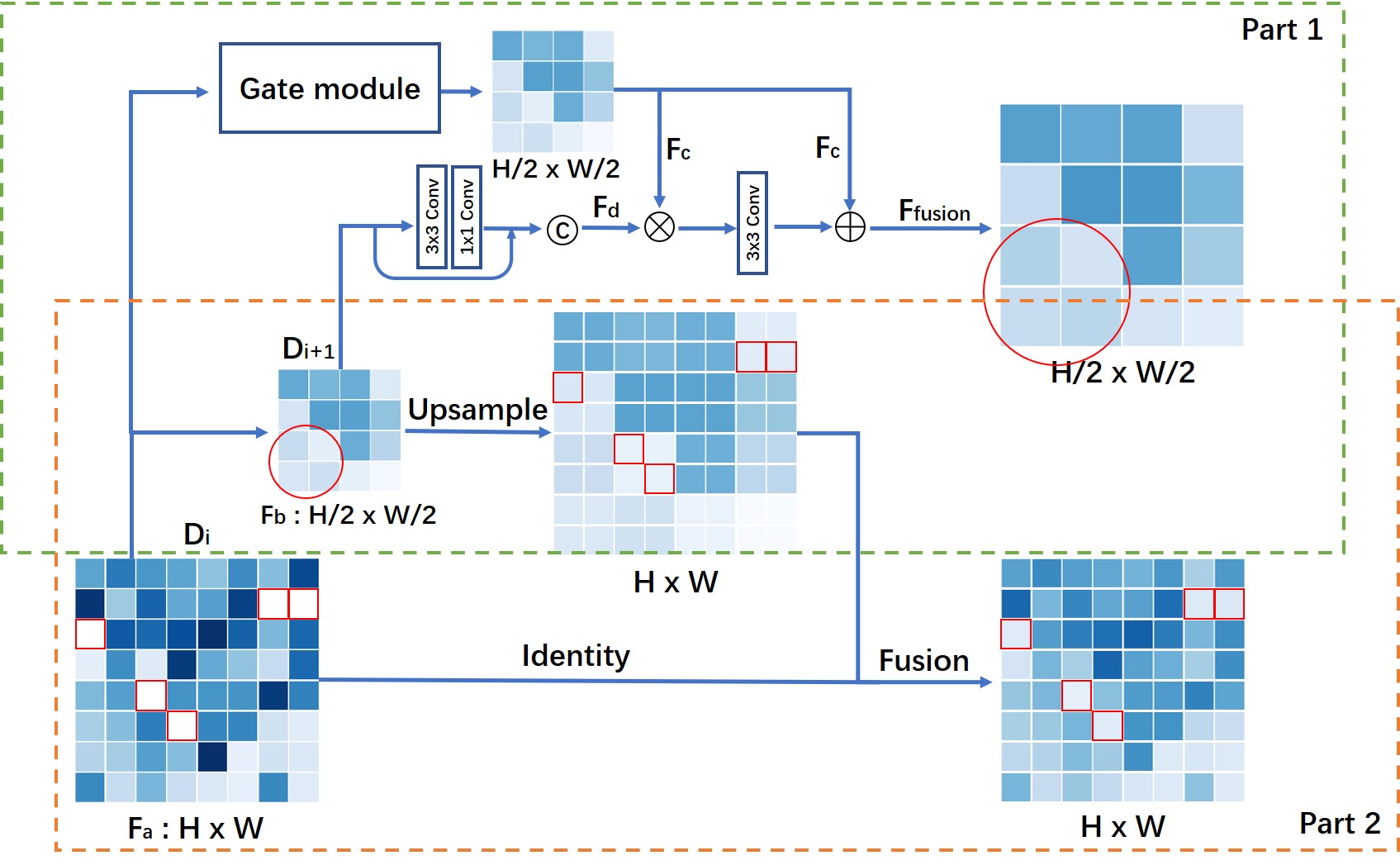}
  \caption{Feature fusion module. Part 1 fuses the features at the same scale, which takes advantage of the complementary information of same scale. Part 2 learns missing points in neural descriptors by fusing the features at different scales.}
    \label{fig:label4}
\end{figure}

\subsubsection{Fusing features at different scales}
The lack of topology information in the point cloud leads to holes and bleeding. Given the feature $F_a$ of the neural descriptor $D_{i}$ with holes, as shown in the red box in Fig.~\ref{fig:label4}. Although $D_{i}$ has a higher resolution with fine details, it still suffers from larger surface bleeding. For feature blocks in $F_a$ with no values, rough values can be obtained after average pooling. $D_{i+1}$ has a low resolution, however, it is able to reduce surface bleeding. For the sparse point cloud, fusing features at two different scales still cannot completely fill the hole. Therefore, we suggest $\omega-net$ to fuse multi-scale features. In our model, we use neural descriptors of four scales in order to achieve the trade-offs between efficiency and accuracy. Through fusing features at different scales, our proposed model learns the missing points in the sparse point cloud from the data, so as to synthesize a realistic novel view of the sky, distant objects, etc. Instead of using transpose convolution, we employ bilinear interpolation in upsampling phase. This is because transpose convolution is basically learnable upsampling, and learnable parameters incur the extra computation.

\subsubsection{Fusing features at the same scale}
In our presented method, the feature $F_b$ of neural descriptor $D_{i+i}$ is obtained from $D_{i}$ by an average pooling operation. The down-sampled feature $F_d$ of neural descriptor $D_{i+i}$ concatenates itself with the last $1 \times 1$ layer feature, which is used for detail enhancement to retain the lost information. At the same time, the feature $F_c$ with the size of $\frac{H}{2} \times \frac{W}{2}$ is obtained by the neural descriptor $D_{i}$ of the gate module using the gate convolution. The fusion of $F_c$ and $F_d$ features at the same scale can make use of complementary information between features, where the fusion feature is $F_{fusion}=F_c+F_d\otimes F_c$, as shown in the red circle of Fig.~\ref{fig:label4}.


\begin{figure}[h]
\setlength{\abovecaptionskip}{-0.1cm} 
\setlength{\belowcaptionskip}{0pt}
  \centering
  \includegraphics[scale=0.4]{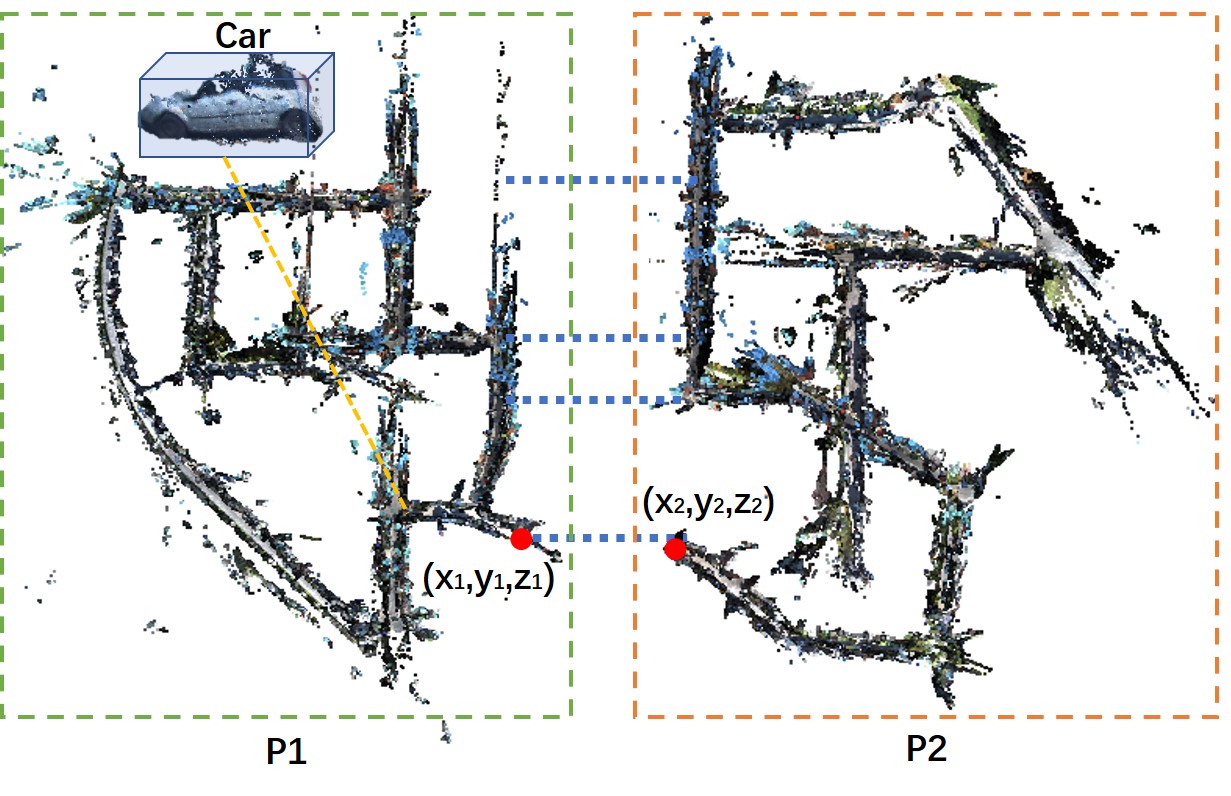}
  \caption{The scene stitching.}
    \label{fig:label5}
    \vspace{-0.3cm}
\end{figure}

\subsection{Scene Editing and Stitching}
As our proposed model learns neural descriptors from point clouds to represent scenes, the scene can be edited by changing the trained neural descriptors. Therefore, we can move the cars, people, and houses in the scene at will. Or even simulate the extreme scenarios, like a car going the wrong way is about to crash. 

As shown in Fig.\ref{fig:label5}, $\{(x,y,z)\mid x\in(x_{min},x_{max}), y\in(y_{min},y_{max}),\\ z\in(z_{min},z_{max})\}$ represent the range of a car in the point clouds. Through back propagation, we employ a rendering network with learnable parameter $\theta$ to project all the neural descriptors $D$ onto the RGB image
\begin{equation}
I_{x, y}=\psi_{\mathrm{pr}}\left(P, D, C, \theta\right), 
\end{equation}
where $I_{x,y}$ is the projected 2D image, and $\psi_{\mathrm{pr}}$ denotes the projection and rasterization process. We synthesize the novel view of car $I_{x', y'}=\psi_{\mathrm{pr}}\left(P', D', C, \theta\right)$ via changing its position of car $p'=(x',y',z')$. By taking advantage of scene editing, we can not only move objects in the scene, but also remove dynamic objects, so as to obtain more diverse driving scenes, as shown in Fig.\ref{fig:label7}.

To account for the large scale driving scene, we propose a scene stitching method that is able to concatenate multiple scenes and update a block locally. For coordinates at the boundary of scene 1 $(x_1,y_1,z_1)$, the boundary coordinates of scene 2 $(x_2,y_2,z_2)$ need to be stitched. We firstly rotate the point clouds $(P_{1},P_{2})$ of the two scenes so that they are aligned on a coordinate system at the boundary. The feature descriptors $(D_{1},D_{2})$ represent the texture of their scenes after being trained by our rendering network. Then, $D_{1}$ and $D_{2}$ are stitched at the boundary to update the scene. The new scene is $concat(D_{1},D_{2})$.

\begin{figure*}[t]
  \centering
  \includegraphics[width=\linewidth]{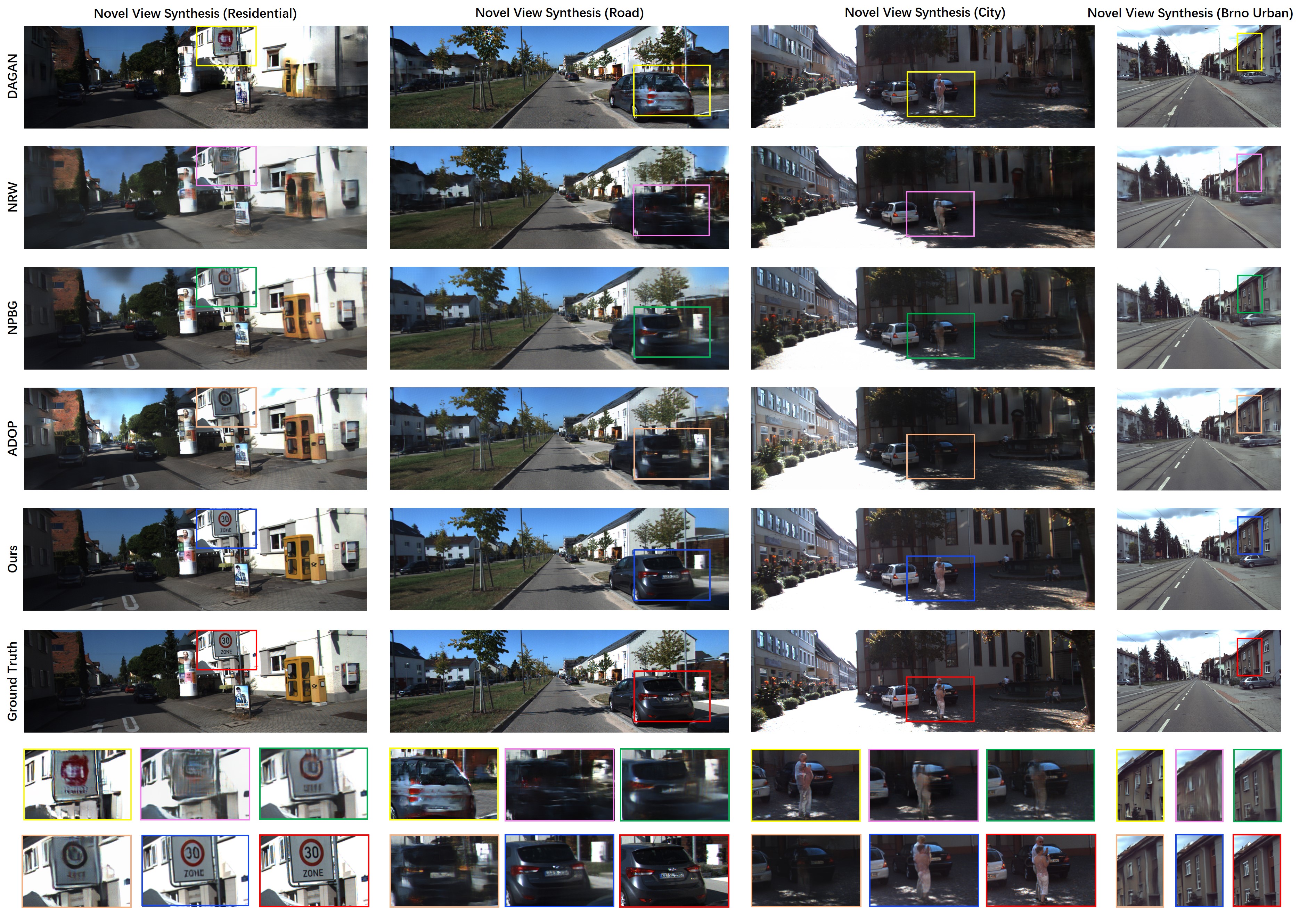}
  \caption{Comparative results of novel view synthesis on the Residential, Road, City scenes from the KITTI benchmark, and a multiple view scene from the Brno Urban dataset. 
  Comparing to DAGAN~\cite{tang2020dual}, NRW~\cite{meshry2019neural}, NPBG~\cite{aliev2020neural} and ADOP~\cite{ruckert2021adop}, our approach performs the best in cases of pedestrians, vehicles, sky, buildings and road signs. Please zoom in for more details.}
   \label{fig:label6}
\end{figure*}

\subsubsection{Loss function}
Perceptual loss~\cite{johnson2016perceptual}, also known as VGG loss, can effectively reflect the image quality of perception more than other loss functions. Thus, we employ the perceptual loss function to prevent smoothing the high-frequency details while encouraging color preservation. Specifically, we compute the perceptual loss between the synthetic novel view and ground truth image $I_{GT}$, which is calculated by a pretrained VGG layer $\Phi_{l}$ as follows:
\begin{equation}
L\left(D, \theta\right)=\sum_{l=1}^L \left(\Phi_{l}\left(I_{GT}\right)-\Phi_{l}\left(R_{\theta}\left(P,D,C \right)\right)\right),
\end{equation}
where $l$ denotes the randomly cropped patches. Given point cloud $P$ and camera parameters $C$, our driving scene render $R_{\theta}$ learns the neural descriptors $D$ and network parameters $\theta$.

\section{Experiment}
To fairly compare the qualitative and quantitative results of various methods, we conduct the experiments on Nvidia GeForce RTX 2070 GPU and evaluate our proposed approach on the two datasets for autonomous driving. To reduce memory usage to load the point cloud of the entire large-scale scene, all comparison methods use the sparse point cloud optimized by our method as input.

\subsection{Datasets}

\pmb{KITTI Dataset}~\cite{geiger2012we}: KITTI is a large data set of real driving scenarios, which contains rich scenes. We mainly conducted experiments in three different cases, namely Residential, Road and City, covering 3724, 996, and 1335 meters, respectively. Due to the overlapping parts, we finally adopted 3560 frames, 819 frames, and 1584 frames in Residential, Road, and City scenes as the testbed. We evaluated every 10 frames (e.g., frame 0, 10, 20...) by following the training and testing split of~\cite{aliev2020neural,ruckert2021adop}. The rest of the frames are used for training. To demonstrate the effectiveness of our method, we conducted the more challenging experiment by discarding 5 test frames before and after every 100 frames as test frames. As the car is driving at a fast speed, losing 10 frames may lose a lot of scene information, especially at corners.


\pmb{Brno Urban Dataset}~\cite{ligocki2020brno}: Compared to KITTI's single-view trajectory, the Brno Urban Dataset contains four views, including left, right, left-front and right-front. In this paper, we use 1641 frames of driving images in our experiments. The left view, the left-front view, and the right view are used as a set of data using similar evaluation criteria as KITTI.


\subsection{Evaluation on the KITTI Testing Set}

Since point clouds from external reconstruction methods such as MetaShape~\cite{Metashape} often contain holes and outliers, the quality of rendering is usually degraded. Moreover, the sparse point cloud as input brings a great challenge to scene synthesis.

To demonstrate the efficacy of our method, we compare it against the recent image-to-image translation based and neural point-based approaches, including DAGAN~\cite{tang2020dual}, NRW~\cite{meshry2019neural}, NPBG~\cite{aliev2020neural} and ADOP~\cite{ruckert2021adop}, which have achieved promising results in outdoor scene synthesis. Followed by the above methods, we employ Peak Signal-to-Noise Ratio (PSNR), Structural Similarity (SSIM) and perceptual loss (VGG loss) as the evaluation metrics. To facilitate the fair comparison, we also adopt the perception metric and learned Perceptual Image Patch Similarity (LPIPS) in our evaluation.


Tang et al.~\cite{tang2020dual} propose a Novel Dual Attention GAN (DAGAN) algorithm, which can effectively model semantic attention at spatial and channel dimensions, thus improving the feature representation ability of semantic images. The GAN-based method can synthesize reasonable semantic images, however, there is a gap with the real scene texture, and its performance is weak in various metrics.

NRW~\cite{meshry2019neural} renders the points to the deep frame buffer and learns the mapping from the initial rendering to the actual photo by training the neural network. Besides point clouds, camera views, and paired poses, NRW needs the extra depth maps for supervision, which are obtained by Agisoft Metashape~\cite{Metashape}. Although having obtained good metrics in KITTI dataset, it does not perform well in the texture with details, as shown in Fig.\ref{fig:label6}. 


NPBG~\cite{aliev2020neural} is one of the baseline methods. To facilitate fair comparisons, we use the same parameters, such as the number of neural descriptors and the number of layers of the network structure, etc. We evaluate the effectiveness of our method in detail by comparing ablation with NPBG.

ADOP~\cite{ruckert2021adop} proposes the differentiable render that can optimize the parameters of camera pose, intrinsic, or texture color. However, it has complicated scene parameters of pipeline and is a two-stage rendering network, per image exposure and per image white balance needs to be manually adjusted. ADOP has difficulty in sparse point clouds, which tends to synthesize the blurred images in the area of point clouds with holes.

In our proposed approach, we initialize the driving scene through point cloud in four datasets and use neural descriptors to encode the geometry and appearance. At each iteration, we sample ten target patches with the size of $256 \times 256$ in KITTI Dataset. Due to the high image resolution of the Brno Urban Dataset, the patch with the size of $336 \times 336$ is used for training. In Monte Carlo sampling, we set the sampling ratio to 80\%. Table~\ref{tab:table1} shows the results of our method. It can be seen that our proposed approach is significantly better than the previous method on all metrics.

\begin{table*}
\centering
\caption{Quantitative evaluation of novel view synthesis on three scenes from the KITTI dataset.}
\begin{tabular}{l|c|c|c}
\toprule
   & KITTI Residential & KITTI Road & KITTI City \\
& VGG$\downarrow$ \enspace PSNR$\uparrow$ \enspace LPIPS$\downarrow$ \enspace  SSIM$\uparrow$
& VGG$\downarrow$ \enspace PSNR$\uparrow$ \enspace LPIPS$\downarrow$ \enspace  SSIM$\uparrow$
& VGG$\downarrow$ \enspace PSNR$\uparrow$ \enspace LPIPS$\downarrow$ \enspace  SSIM$\uparrow$  \\
\midrule
\multicolumn{4}{c}{Test every 100 frames (w/ discard)} \\
\midrule
DAGAN~\cite{tang2020dual} &   1241.3 \! \enspace 11.18 \enspace \enspace 0.4968 \enspace \enspace 0.3081  & 929.0 \enspace \enspace 15.33 \enspace \enspace 0.3570 \enspace \enspace 0.4135  & 1301.3 \! \enspace 10.74 \enspace \enspace 0.4949 \enspace \enspace 0.3014  \\
NRW~\cite{meshry2019neural} &  923.0 \enspace \enspace 15.70 \enspace \enspace 0.3874 \enspace \enspace 0.4748  & 860.8 \enspace \enspace 17.01 \enspace \enspace 0.3343 \enspace \enspace 0.4311  & 1007.0 \! \enspace 15.66 \enspace \enspace 0.3847 \enspace \enspace 0.4361    \\
NPBG~\cite{aliev2020neural} & 924.7 \enspace \enspace 14.98 \enspace \enspace 0.4426 \enspace \enspace 0.4733  & 791.4 \enspace \enspace 17.63 \enspace \enspace 0.3680 \enspace \enspace 0.5080  & 994.5 \enspace \enspace 14.97 \enspace \enspace 0.4384 \enspace \enspace 0.4518 \\
ADOP~\cite{ruckert2021adop} &  900.5 \enspace \enspace 14.89 \enspace \enspace 0.3590 \enspace \enspace 0.4734  & 785.9 \enspace \enspace 17.56 \enspace \enspace 0.3275 \enspace \enspace 0.4701  & 910.6 \enspace \enspace 15.67 \enspace \enspace 0.3497 \enspace \enspace 0.4774  \\

READ~(Ours)  & \pmb{695.3} \enspace \enspace \pmb{17.70} \enspace \enspace \pmb{0.2875} \enspace \enspace \pmb{0.5963} & \pmb{573.5} \enspace \enspace \pmb{20.26} \enspace \enspace \pmb{0.2408} \enspace \enspace \pmb{0.6238} & \pmb{673.2} \enspace \enspace \pmb{18.35} \enspace \enspace \pmb{0.2529} \enspace \enspace \pmb{0.6412} \\ 
\midrule
\multicolumn{4}{c}{Test every 10 frames (w/o discard)} \\
\midrule
DAGAN~\cite{tang2020dual}  & 1031.2 \! \enspace 14.27 \enspace \enspace 0.3800 \enspace \enspace 0.4337 & 847.2 \enspace \enspace 16.84 \enspace \enspace 0.2916 \enspace \enspace 0.4638 & 1128.8 \! \enspace 13.40 \enspace \enspace 0.3971 \enspace \enspace 0.3845 \\
NRW~\cite{meshry2019neural} & 767.4 \enspace \enspace 18.43 \enspace \enspace 0.3197 \enspace \enspace 0.5476 & 748.0 \enspace \enspace 18.58 \enspace \enspace 0.2809 \enspace \enspace 0.4996 & 823.7 \enspace \enspace 18.02 \enspace \enspace 0.3102 \enspace \enspace 0.5682 \\
NPBG~\cite{aliev2020neural} & 621.2 \enspace \enspace 19.32 \enspace \enspace 0.2584 \enspace \enspace 0.6316  & 597.3 \enspace \enspace 20.25 \enspace \enspace 0.2517 \enspace \enspace 0.5919  & 632.8 \enspace \enspace 19.58 \enspace \enspace 0.2480 \enspace \enspace 0.6277 \\
ADOP~\cite{ruckert2021adop} & 610.8 \enspace \enspace 19.07 \enspace \enspace 0.2116 \enspace \enspace 0.5659 & 577.7 \enspace \enspace 19.67 \enspace \enspace 0.2150 \enspace \enspace 0.5554 & 560.9 \enspace \enspace 20.08 \enspace \enspace 0.1825 \enspace \enspace 0.6234 \\
READ~(Ours)  & \pmb{454.9} \enspace \enspace \pmb{22.09} \enspace \enspace \pmb{0.1755} \enspace \enspace \pmb{0.7242} & \pmb{368.2} \enspace \enspace \pmb{24.29} \enspace \enspace \pmb{0.1465} \enspace \enspace \pmb{0.7402} & \pmb{391.1} \enspace \enspace \pmb{23.48} \enspace \enspace \pmb{0.1321} \enspace \enspace \pmb{0.7871}

\\

\bottomrule
\end{tabular}

\label{tab:table1}
\end{table*}

\begin{table*}
\centering
\caption{Quantitative evaluation of novel view synthesis on Brno Urban Dataset.}
\begin{tabular}{l|c|c|c|c}
\toprule
   & Left side view & Left front side view & Right side view & Total \\
& VGG$\downarrow$ \enspace PSNR$\uparrow$ \enspace LPIPS$\downarrow$ \enspace  SSIM$\uparrow$
& VGG$\downarrow$ \enspace PSNR$\uparrow$ \enspace LPIPS$\downarrow$ \enspace  SSIM$\uparrow$
& VGG$\downarrow$ \enspace PSNR$\uparrow$ \enspace LPIPS$\downarrow$ \enspace  SSIM$\uparrow$ 
&   PSNR$\uparrow$ \enspace LPIPS$\downarrow$ 
\\
\midrule
\multicolumn{5}{c}{Test every 100 frames(w/ discard)} \\
\midrule
DAGAN~\cite{tang2020dual} &  1055.5 \! \enspace 13.93 \enspace \enspace \pmb{0.3960} \enspace \enspace 0.3705  & 754.9 \enspace \enspace 16.95 \enspace \enspace 0.3078 \enspace \enspace 0.5234  & 1105.1 \! \enspace 11.84 \enspace \enspace \pmb{0.5323} \enspace \enspace 0.3561   & 14.24 \enspace \enspace 0.4120  \\
NRW~\cite{meshry2019neural} &  919.2 \enspace \enspace 15.25 \enspace \enspace 0.4435 \enspace \enspace 0.4397  & 712.7 \enspace \enspace 17.87 \enspace \enspace 0.4063 \enspace \enspace 0.5513  & 949.5 \enspace \enspace 13.49 \enspace \enspace 0.5790 \enspace \enspace 0.4405    & 15.54 \enspace \enspace 0.4762   \\
NPBG~\cite{aliev2020neural} & 1002.3 \! \enspace 13.14 \enspace \enspace 0.5242 \enspace \enspace 0.3978  & 724.5 \enspace \enspace 17.13 \enspace \enspace 0.4098 \enspace \enspace 0.5596  & 1024.4 \! \enspace 12.22 \enspace \enspace 0.6634 \enspace \enspace 0.4333 & 14.17 \enspace \enspace 0.5325 \\
ADOP~\cite{ruckert2021adop} &   997.1 \enspace \enspace 14.08 \enspace \enspace 0.4373 \enspace \enspace 0.3915  & 683.6 \enspace \enspace 18.24 \enspace \enspace 0.3150 \enspace \enspace 0.5618  & 1091.2 \! \enspace 13.21 \enspace \enspace 0.5531 \enspace \enspace 0.3803  & 15.18 \enspace \enspace 0.4352  \\
READ (ours)  & \pmb{842.0} \enspace \enspace \pmb{15.28} \enspace \enspace 0.3992 \enspace \enspace \pmb{0.4752} & \pmb{523.9} \enspace \enspace \pmb{20.51} \enspace \enspace \pmb{0.2467} \enspace \enspace \pmb{0.6713} & \pmb{928.0} \enspace \enspace \pmb{13.88} \enspace \enspace 0.5464 \enspace \enspace \pmb{0.4533} & \pmb{16.56} \enspace \enspace \pmb{0.3974}\\

\midrule
\multicolumn{5}{c}{Test every 10 frames(w/o discard)} \\
\midrule
DAGAN~\cite{tang2020dual} & 851.4 \enspace \enspace 16.67 \enspace \enspace 0.2822 \enspace \enspace 0.4766 & 657.6 \enspace \enspace 19.08 \enspace \enspace 0.2445 \enspace \enspace 0.5662 & 1041.6 \! \enspace 13.14 \enspace \enspace 0.4514 \enspace \enspace 0.3805 &  16.30  \enspace \enspace 0.3260  \\
NRW~\cite{meshry2019neural} & 735.0 \enspace \enspace 18.64 \enspace \enspace 0.3199 \enspace \enspace 0.5422 & 619.6 \enspace \enspace 19.74 \enspace \enspace 0.3125 \enspace \enspace 0.6062 & 864.6 \enspace \enspace 16.05 \enspace \enspace 0.4631 \enspace \enspace 0.4749 &  18.14 \enspace \enspace 0.3651 \\
NPBG~\cite{aliev2020neural} & 659.4 \enspace \enspace 18.56 \enspace \enspace 0.3112 \enspace \enspace 0.5849  & 531.6 \enspace \enspace 20.30 \enspace \enspace 0.2705 \enspace \enspace 0.6773  & 813.1 \enspace \enspace 16.00 \enspace \enspace 0.4424 \enspace \enspace 0.5093 &  18.28 \enspace \enspace 0.3414  \\
ADOP~\cite{ruckert2021adop} & 634.0 \enspace \enspace 19.19 \enspace \enspace 0.2414 \enspace \enspace 0.5927 & 520.6 \enspace \enspace 20.83 \enspace \enspace 0.2189 \enspace \enspace 0.6633 & 807.1 \enspace \enspace 16.51 \enspace \enspace 0.3636 \enspace \enspace 0.5009 &  18.84 \enspace \enspace 0.2746 \\
READ (ours)  & \pmb{459.8} \enspace \enspace \pmb{21.79} \enspace \enspace \pmb{0.1905} \enspace \enspace \pmb{0.7067} & \pmb{341.1} \enspace \enspace \pmb{24.85} \enspace \enspace \pmb{0.1513} \enspace \enspace \pmb{0.7836} & \pmb{663.6} \enspace \enspace \pmb{18.44} \enspace \enspace \pmb{0.3065} \enspace \enspace \pmb{0.5771}  & \pmb{21.69}  \enspace \enspace \pmb{0.2161}  \\
	
\bottomrule
\end{tabular}

\label{tab:table2}
\end{table*}

\subsection{Evaluation on the Brno Urban Testing Set}
Unlike the KITTI dataset, the Brno Urban dataset is very challenging with three views.

In the evaluation, we test the left side view, left front side view, and right side view, separately. As shown in Table~\ref{tab:table2}, our proposed approach is significantly better than the other methods while DAGAN obtains slightly better LPIPS metrics in the 100 frame test. The results of NRW's side view are similar to ours. This is due to the faster car speed and the narrower side view, resulting in significantly different images between frames. As the distributions of training and test sets are different, the methods based on point cloud rendering are seriously affected. NRW relies on additional segmentation annotations to synthesize the unfamiliar scenes, and the image is projected onto the point cloud to initialize the deep buffer. The method using the GAN generator can obtain reasonable images for the missing regions of the point cloud. Therefore, DAGAN and NRW achieve a slight improvement in the case of fewer input images. However, such methods are limited by synthesizing novel scenes from different views. In addition, ADOP uses a differentiable rendering pipeline to align camera images to point clouds, the right side view synthesized results are similar to ours.


\subsection{Ablation Study}
In the ablation experiment, we examine the effectiveness of each module, and more results are given in the supplementary materials. For a fair comparison, we have added the sampling strategy proposed in Section 3.2.3 to the NPBG as our baseline, namely sampling NPBG, as shown at the first line in Table~\ref{tab:table3}. Then, we gradually add each module mentioned in Section~\ref{sec:LNSR} and evaluate them in the KITTI Road scenario. In contrast to sampling NPBG, our proposed basic gate module can effectively filter the invalid values in neural descriptors, which obtains significant improvement in PSNR, SSIM, and other metrics. By fusing the features at the same scale~(Same), the texture of the scene is enhanced with fine details. The fusion of different scale features modules~(Differ) can effectively fill the value of neural descriptors close to zero. All metrics are greatly improved over the baseline. We also study the influence of different loss functions. It can be observed that combining ${L_1}$, PSNR Loss $L_{PSNR}$ and VGG Loss can improve the SSIM index slightly.


\begin{table}[th]
\caption{Ablation study of our method on KITTI road dataset.}
\begin{tabular}{ c c c c c | c c c c }
\hline

\rotatebox{90}{Gate} & \rotatebox{90}{Same} & \rotatebox{90}{Differ} & \rotatebox{90}{$L_{1}$} & \rotatebox{90}{$L_{PSNR}$}  & VGG$\downarrow$ &   PSNR$\uparrow$ & LPIPS $\downarrow$ & SSIM$\uparrow$ 
\\

\hline 

\quad &\quad &\quad &\quad &\quad  & 572.1 & 20.63 & 0.2359 & 0.6109   \\
\Checkmark &\quad &\quad &\quad &\quad  &  477.6 & 22.29  & 0.1893 & 0.6883 \\
\Checkmark &\Checkmark &\quad &\quad &\quad  & 392.1 & 23.76  & 0.1477 & 0.7205  \\
\Checkmark &\Checkmark &\Checkmark  &\quad &\quad  & \pmb{368.2} &   \pmb{24.29}  & \pmb{0.1465} & 0.7402  \\
\hline 
\Checkmark &\Checkmark &\Checkmark &\Checkmark &\quad  &  401.3 & 24.16 & 0.1865 & 0.7487 \\
\Checkmark &\Checkmark &\Checkmark &\quad &\Checkmark  &  383.3 & 23.96 & 0.1506 & 0.7325 \\
\Checkmark &\Checkmark &\Checkmark &\Checkmark &\Checkmark &   401.1 & 24.19 & 0.1863 & \pmb{0.7490} \\ 
                     
\hline
\end{tabular}

\label{tab:table3}
\end{table}

\begin{table}[h]
\caption{Comparisons of scene stitching on KITTI  dataset.}
\centering
\begin{tabular}{l|c}
\toprule
Method  & VGG$\downarrow$ \enspace   PSNR$\uparrow$ \enspace LPIPS$\downarrow$ \enspace SSIM$\uparrow$ \\
\midrule
READ &  454.9 \enspace \enspace  22.08 \enspace \enspace 0.1755  \enspace \enspace 0.7242  \\
READ w/ stitching &  429.3 \enspace \enspace  22.58 \enspace \enspace 0.1625  \enspace \enspace 0.7392  \\
\bottomrule
\end{tabular}

\label{tab:table4}
\end{table}

\subsection{Driving Scene Editing}
Editing the driving scenarios not only provides more synthetic data for Advanced Driver Assistance Systems, but also simulates the rare traffic conditions in daily life, i.e., a car driving illegally on the wrong side of the road. Moreover, our proposed approach can remove the dynamic objects in the scene so that data collection staff do not need to worry about the impact of complex traffic and vehicles on the restoration of the real scene. This provides convenience for data collection. Additionally, the panoramic view can be synthesized through our method. The larger field-of-view provides more street view information for Driver Assistance Systems, which makes it easier to observe the surrounding environment and deal with the emergencies in a timely manner, as shown in Fig.~\ref{fig:label7}. More results are presented in the supplementary materials.

\begin{figure}[h]
  \centering
  \includegraphics[width=\linewidth]{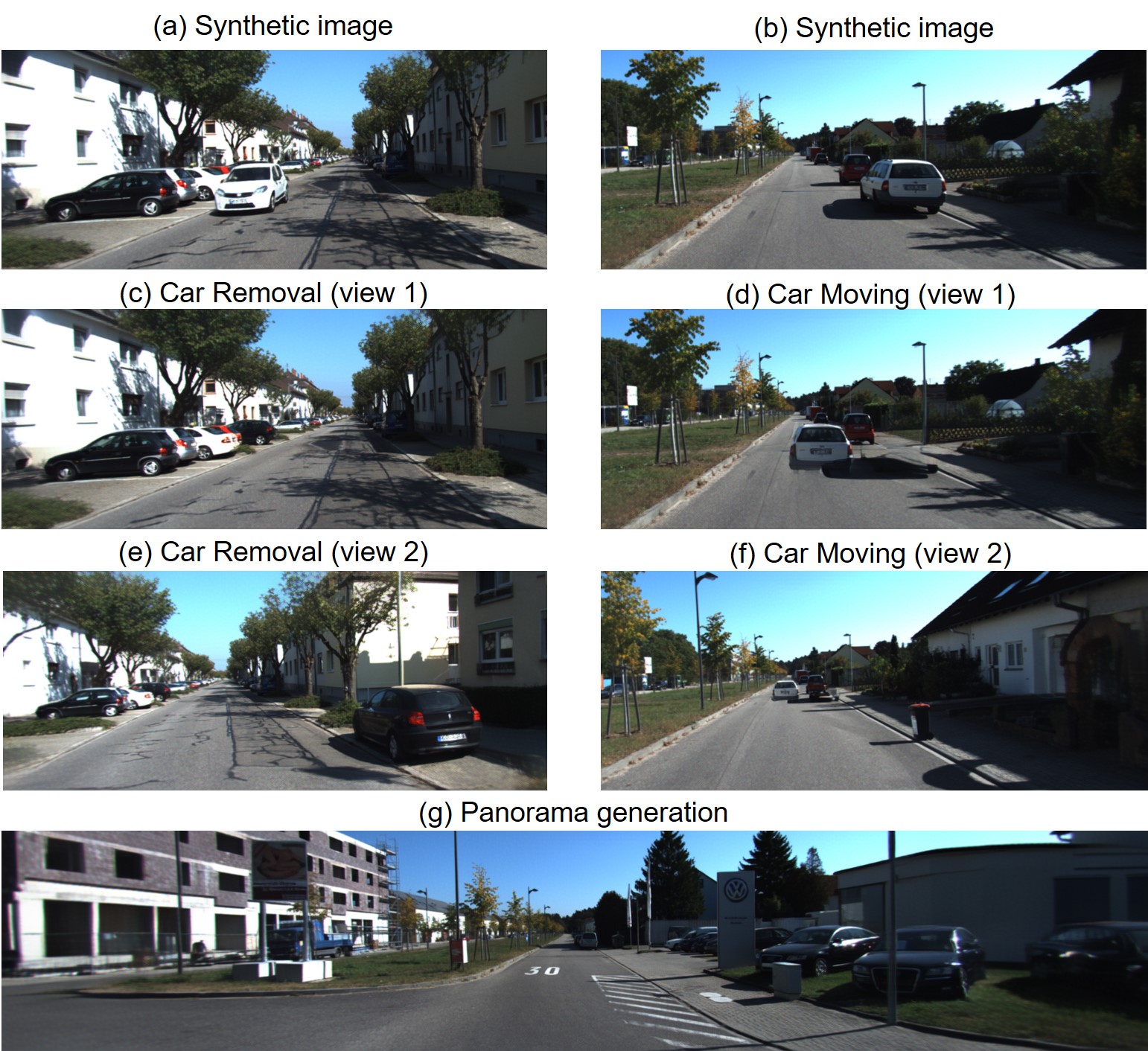}
  \caption{Example results of scene editing. We can move and remove the cars in different views. A panorama with larger view can be synthesized by changing the camera parameters.}
\label{fig:label7}
\end{figure}
\subsection{Driving Scene Stitching}
By taking advantage of scene stitching, our model is able to synthesize the larger driving scenes and update local areas with obvious changes in road conditions. This not only enables to deal with the driving area at a larger scale, but also divides the large-scale scene into small parts for efficient training in parallel. As shown in Fig.~\ref{fig:label5}, we stitch two KITTI residential scenes, which share the same rendering network and learn the texture from the corresponding part, respectively. Table~\ref{tab:table4} shows the comparison results on the stitched scenario. It can be seen that decomposing large scenes into small parts can achieve better results. This indicates the effectiveness of our presented stitching method.
 

\section{Limitations}
\begin{figure}[h]
  \centering
  \includegraphics[width=\linewidth]{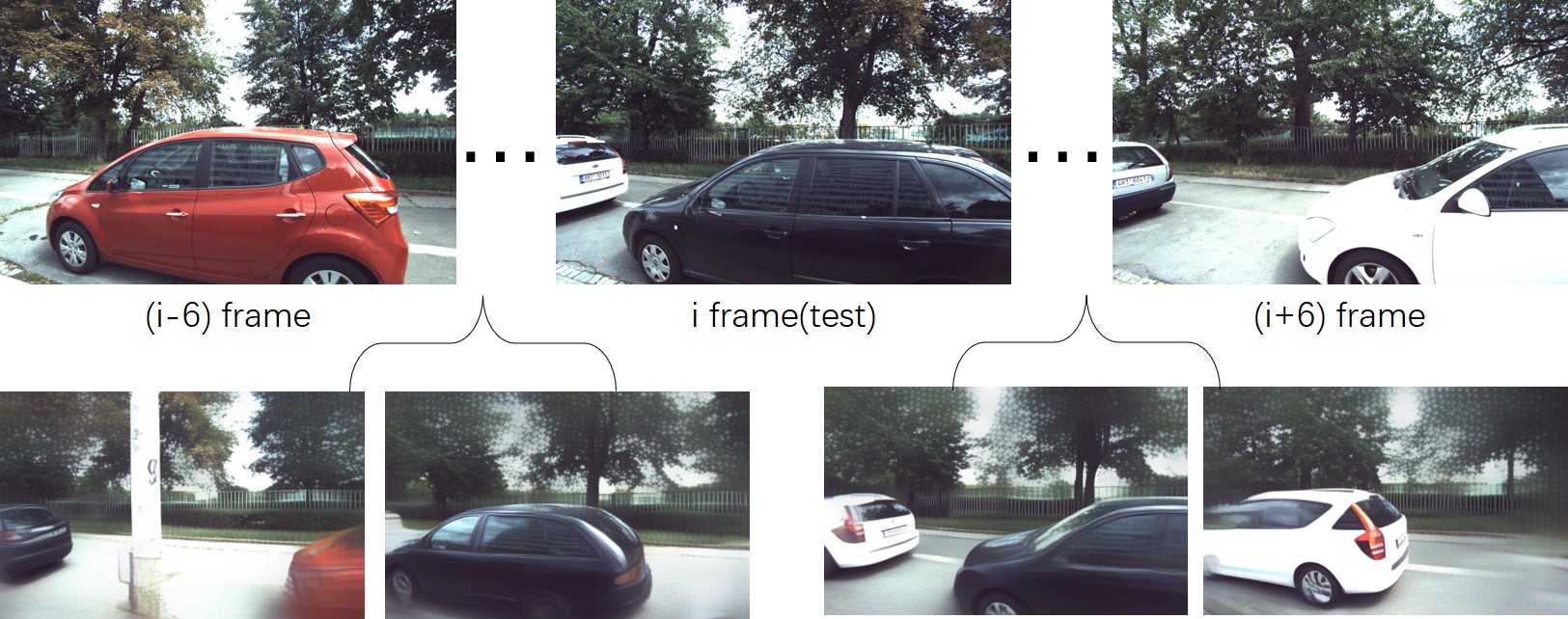}
  \caption{Failure cases.}
\label{fig:label8}
\end{figure}

We propose a multi-scale rendering method that synthesizes the photo-realistic driving scenes from sparse point clouds. However, for images that differ greatly from the training views, for example, in the right view of the Brno Urban data set, 10 frames near the test frame are discarded. As shown in Fig.\ref{fig:label8}. neural rendering-based methods are difficult to synthesize the scene with few observations, resulting in blur. In addition, misalignment on point clouds affects the rendering results. In the future, we will consider using point clouds scanned by LiDAR sensor as training data so as to reduce the reconstruction errors.

\section{Conclusion}
This paper proposed an efficient neural scene rendering approach to autonomous driving, which makes it possible to synthesize large-scale scenarios on a PC through a variety of sampling schemes. We presented an $\omega-net$ rendering network to filter the neural descriptors through basic gate modules, and fused features at the same scale and different scales with different strategies. Our proposed approach not only synthesized the photo-realistic views, but also edited and stitched the driving scenes. This enables to generate various photo-realistic images to train and test the autonomous driving system. The encouraging experimental results showed that our proposed approach significantly outperforms the alternative methods both qualitatively and quantitatively.


\newpage

\bibliographystyle{ACM-Reference-Format}
\bibliography{sample-sigconf}

\appendix

\end{document}